\documentclass[letterpaper, 10 pt, conference]{ieeeconf}  

\IEEEoverridecommandlockouts                              

\usepackage{amsmath}
\usepackage{amssymb}
\usepackage{hyperref}
\usepackage{multirow}
\usepackage{graphicx}

\title{\LARGE \bf
4DR P2T: 4D Radar Tensor Synthesis with Point Clouds
}

\author{Woo-Jin Jung, Dong-Hee Paek, and Seung-Hyun Kong
\thanks{This work was supported by the National Research Foundation of Korea(NRF) grant funded by the Korea government(MSIT) (No. 2021R1A2C3008370).}
\thanks{Woo-Jin Jung, Dong-Hee Paek, and Seung-Hyun Kong are with the CCS Graduate School of Mobility, Korea Advanced Institute of Science and Technology, Daejeon, Korea, 34051 
        {\tt\small \{woo-jin.jung, donghee.paek, skong\}@kaist.ac.kr}}}%

\begin{document}

\maketitle
\thispagestyle{empty}
\pagestyle{empty}

\begin{abstract}

In four-dimensional (4D) Radar-based point cloud generation, clutter removal is commonly performed using the constant false alarm rate (CFAR) algorithm. However, CFAR may not fully capture the spatial characteristics of objects. To address limitation, this paper proposes the 4D Radar Point-to-Tensor (4DR P2T) model, which generates tensor data suitable for deep learning applications while minimizing measurement loss. Our method employs a conditional generative adversarial network (cGAN), modified to effectively process 4D Radar point cloud data and generate tensor data. Experimental results on the K-Radar dataset validate the effectiveness of the 4DR P2T model, achieving an average PSNR of 30.39dB and SSIM of 0.96. Additionally, our analysis of different point cloud generation methods highlights that the 5\% percentile method provides the best overall performance, while the 1\% percentile method optimally balances data volume reduction and performance, making it well-suited for deep learning applications.

\end{abstract}

\section{INTRODUCTION}

In recent autonomous driving research, 4D Radar has gained increasing attention as an advanced sensing technology. Traditional Radar sensors, often employed as auxiliary sensors, measure range, azimuth, and Doppler information. In contrast, 4D Radar incorporates elevation into these measurements, enabling more precise spatial perception. Consequently, 4D Radar provides more robust measurements than cameras and LiDAR under adverse weather conditions such as snow or rain. Furthermore, it surpasses traditional Radar in detecting object contours, demonstrating superior object detection capabilities. Owing to these advantages, 4D Radar has emerged as a key sensing modality in autonomous driving systems, offering enhanced object detection across diverse operational environments.

Most 4D Radar data are provided as point clouds, which are typically generated by traditional handcrafted methods such as CFAR to remove clutter \cite{clutter} from the tensor data. However, CFAR processes each cell independently, disregarding spatial continuity across adjacent cells. As a result, CFAR-generated point clouds often fail to preserve essential spatial characteristics—such as object size, shape, and continuous contours—thereby limiting their ability to accurately represent complex objects \cite{4D_radar_survey}. In autonomous driving scenarios where objects vary in size and shape, this limitation constrains environmental perception. Moreover, CFAR-based point clouds typically exhibit much lower point density than LiDAR, reducing the fidelity of captured object features \cite{rpfa_net} and complicating subsequent sensor fusion processes \cite{dpft}.

\begin{figure}[t!]
  \centering
  \includegraphics[width=1.0\columnwidth]{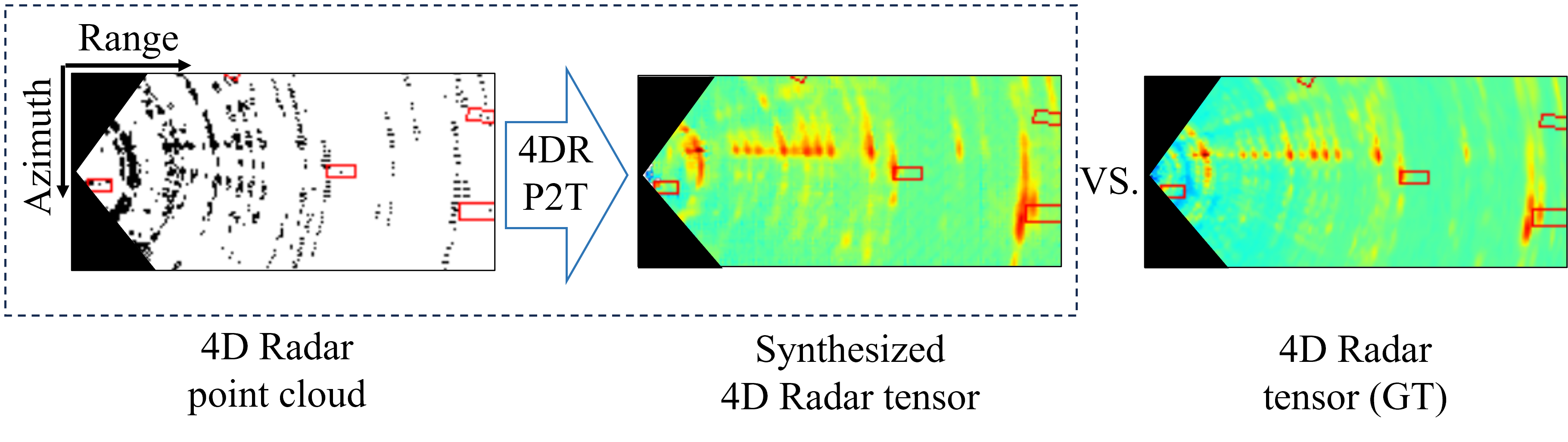}
  \caption{
   4DR P2T overview. The 4DR P2T model generates tensor data from 4D Radar point clouds, which are represented in bird’s-eye view (BEV) as a 2D projection. Traditional point cloud generation methods often suffer from measurement loss, which may affect their suitability for deep learning training. To mitigate this limitation, the model generates tensor data to prevent measurement loss, ensuring that crucial information is retained for deep learning tasks.}
  \label{fig0.overview}
\end{figure}

\begin{figure*}[!th]
 \centering
\vspace{1mm} 
 \includegraphics[width=1.0\textwidth]{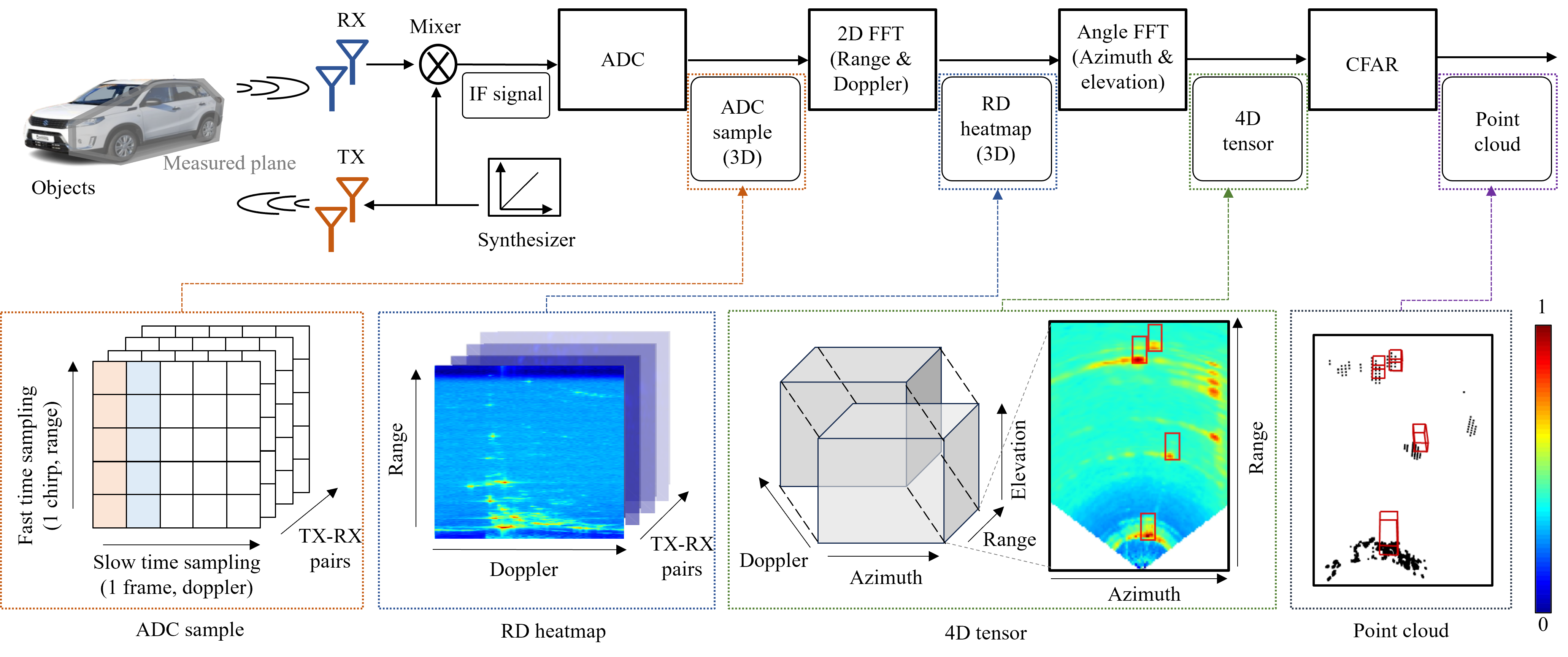}
    \caption{4D Radar signal processing and data representation \cite{4D_radar_survey, 4dradar_tutorial, 4dradar_data_representation}. The Radar power values are normalized and represented using colors. The Radar point cloud is shown as black points, and the bounding box for the objects is indicated with a red box.}
  \label{fig1.4d_radar_data}

\end{figure*}

To mitigate these limitations, previous studies have proposed methods for reconstructing points representing objects \cite{3DRIMR} or generating tensor data prior to CFAR \cite{radarpointgenerator1} \cite{4D_radar_survey}. One notable method \cite{radarpointgenerator1} utilizes a conditional generative adversarial network (cGAN) with a UNet \cite{unet} architecture, leveraging LiDAR data to supervise the generation of denser Radar point clouds. However, fundamental differences between LiDAR (near-infrared) and Radar (electromagnetic waves) result in heterogeneous data characteristics, leading to distortions in power values and contour representations, which may degrade the reliability of the generated Radar data.

As shown in Fig. \ref{fig0.overview}, a method is required to directly generate tensor data using the original 4D Radar tensor as supervision, thereby avoiding cross-sensor inconsistencies. In this study, we leverage the K-Radar dataset \cite{KRadar}, currently the only publicly available dataset that provides 4D tensor data. Prior to its release in 2023, no dataset included 4D tensor data, making direct data-driven methods infeasible. With this new dataset, it is now possible to train models that generate tensor representations from 4D Radar point clouds collected by the same sensor.

Accordingly, we propose the 4D Radar Point cloud-to-Tensor (4DR P2T) model, which utilizes a cGAN-based architecture to generate tensor data from 4D Radar point clouds. This study conducts two primary investigations. First, we identify the point cloud generation method that achieves the best tensor generation performance—measured by peak signal-to-noise ratio (PSNR) and structural similarity index measure (SSIM)—among CFAR \cite{rtnh+} and percentile-based methods \cite{KRadar, enhancekradar} with different densities. Second, we determine the optimal point cloud generation method for deep learning applications, specifically the one that minimizes data volume while preserving sufficiently high tensor generation performance. To enable these investigations, we interpret point cloud data as the encoded version of tensor data, with our 4DR P2T model serving as a decoder that generates the original tensor. Consequently, the tensor generation performance of the 4DR P2T model serves as a proxy for assessing how well a given point cloud preserves environmental information, which in turn facilitates the selection of the most suitable point cloud generation method for deep learning model training and interpretation. Through our experiments, the proposed 4DR P2T model achieves an average PSNR of 30.39dB and SSIM of 0.96, demonstrating its effectiveness and stability. Our findings reveal that the percentile 5\% method yields the best tensor generation performance, while the percentile 1\% method offers an optimal balance between data volume reduction and performance, making it well-suited for deep learning training.

The key contributions of this study are summarized as follows:
\begin{itemize}
    \item Development of the 4DR P2T model, which generates tensor data from 4D Radar point cloud data.
    \item Experimental validation showing that the percentile 5\% data provides the best tensor generation performance.
    \item Confirmation that the percentile 1\% method effectively reduces data volume while maintaining high tensor generation performance.
\end{itemize}

This paper is organized as follows. Section \ref{sec:related_works} discusses 4D Radar signal processing and data generation processes, and reviews related models that convert point cloud data into tensors. Section \ref{sec:method} describes the proposed model architecture. Section \ref{sec:experiments} presents and analyzes the quantitative and qualitative experimental results. Finally, Section \ref{sec:conclusion} concludes the paper and discusses future research directions.

\section{Related work} \label{sec:related_works}
In this section, we provide an overview of related works, focusing specifically on 4D Radar signal processing and data generation, as well as previous studies on data translation methods using cGANs, which form the basis for developing models that generate tensor data from point cloud data.

\begin{figure*}[!th]
    \centering
    \vspace{1mm} 
    \includegraphics[width=1\textwidth]{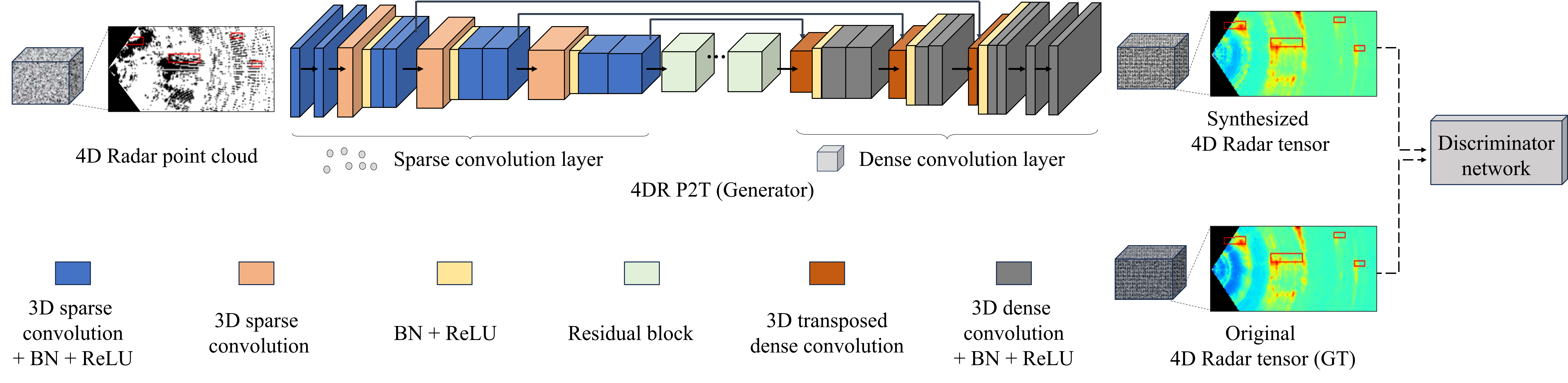} 
    \caption{Overall structure of 4DR P2T. The encoder utilizes 3D sparse convolution to process 4D Radar point cloud data, while the decoder employs 3D dense convolution to generate tensor data.}
    \label{fig2.model}
\end{figure*}

\subsection{4D Radar signal processing and data generation}
The 4D Radar signal processing and data generation process, as applied in autonomous driving, is illustrated in Figure \ref{fig1.4d_radar_data}. The core analog components of a 4D Radar system consist of a synthesizer, transmission (TX) antennas, reception (RX) antennas, and a mixer. The TX antennas emit electromagnetic waves, which reflect off objects in the environment and are received by the RX antennas. The transmitted signal is generated by the synthesizer and radiated through the TX antennas. This signal is a frequency-modulated continuous wave (FMCW), composed of a sequence of frequency-modulated signals, commonly referred to as chirps.

The signal emitted by the TX antennas and the signal received by the RX antennas are combined using a mixer, producing an intermediate frequency (IF) signal. This IF signal represents the frequency difference between the transmitted and received signals, which is used to extract the distance and velocity of the reflected objects. The generated IF signal is then converted into a digital form through an analog-to-digital converter (ADC), creating ADC sample data. This data is separated into a fast time axis, which calculates range information through chirp sampling, and a slow time axis, which calculates Doppler information through frame sampling.

The ADC sample data is processed through a 2D Fast Fourier Transform (FFT), which is applied to perform range FFT and Doppler FFT. The range FFT estimates the distance to objects, while the Doppler FFT estimates their relative velocity, resulting in the generation of an RD heatmap. Although the RD heatmap contains information about range and velocity, it does not include azimuth or elevation information, making it less intuitive to interpret.

To extract azimuth and elevation information, an additional angle FFT is applied to the RD heatmap. The angle FFT utilizes the positional information of the TX and RX antennas arrays in a multiple-input and multiple-output (MIMO) antennas design to analyze the phase differences in the reflected signals. This process generates a 4D tensor that includes range, azimuth, elevation, and Doppler information, with each tensor cell representing the corresponding signal strength. The 4D tensor is represented in a polar coordinate system, but for better interpretability, the visualization shown in the Fig. \ref{fig1.4d_radar_data} is converted into a Cartesian coordinate system.

The generated 4D tensor data is filtered using the CFAR method. CFAR dynamically adjusts the threshold by comparing the signal strength of each cell to its surrounding cells, effectively removing noise and identifying actual targets. This filtering process is applied across all dimensions of the tensor, ultimately producing point cloud data that contains information about actual targets.

The resulting point cloud data includes the position (range, azimuth, elevation) and Doppler of the detected objects and is utilized in various autonomous driving applications, such as object detection and tracking \cite{4dradar_tutorial, 4D_radar_survey, FFT-RadNet}.

\subsection{Image translation}

Image translation focuses on style translation while preserving key information. Notable methods include pix2pix \cite{pix2pix} and pix2pixHD \cite{pix2pixhd}. These methods utilize cGANs to translate input images into output images. These methods have been successfully applied to various image synthesis and transformation tasks \cite{radsimreal, l2r, radarpointgenerator1}. Pix2pix employs a U-Net-based generator and a patch-based discriminator, enabling applications such as image synthesis and color translation. Pix2pixHD extends this framework to handle high-resolution images by incorporating boundary maps, multi-scale generators, and multi-scale discriminators, achieving improved quality. These methods excel at 2D image-to-image translation while maintaining structural information.

However, this study deals with generating tensor data from 4D Radar point cloud data, making it challenging to directly apply conventional image translation models. Existing methods are primarily optimized for 2D image data, necessitating structural modifications to handle higher-dimensional data such as point clouds and tensors. To address this, this study extends the fundamental method of pix2pixHD by modifying the model architecture to effectively process 3D or higher-dimensional data.

\section{Method} \label{sec:method}

This section outlines the data dimensions used for training, the model architecture, and the objective function.

\subsection{Data preparation}
In this study, the 4D Radar tensor data is reduced to 3D spatial information by excluding Doppler information for the training process. As a result, the input data for training consists of four channels, including \(x\), \(y\), \(z\) coordinates, and power values. Including Doppler data would require processing additional values beyond the existing spatial information \((x, y, z)\) and power, necessitating the use of convolution layers with at least four dimensions. This would significantly increase the complexity of the model and computational costs, making it challenging to achieve the primary goal of verifying implementation feasibility in the initial stage of the research. Moreover, according to RTNH \cite{KRadar}, an early model utilizing 4D Radar tensor data, excluding Doppler information still achieves sufficient object detection performance. Therefore, this study focuses on minimizing model complexity while verifying the feasibility of generating tensor data from 4D Radar point cloud data. This method also lays the foundation for future studies incorporating Doppler information.

\subsection{Model structure}
The proposed model is inspired by image translation methods, such as pix2pix and pix2pixHD, and referenced recent studies like l2r \cite{l2r} and RadSimReal \cite{radsimreal} to balance generative feasibility and structural simplicity, while optimizing for input and output data dimensions. To capture the spatial characteristics of Radar point cloud data, the model employs an encoding method based on Voxelnet \cite{voxelnet}, drawing from Lee’s method \cite{l2rtranslation_voxel}. The 4DR P2T model extends the U-Net structure \cite{unet}, commonly used in image translation tasks, with modifications to process 3D data.

As illustrated in Figure \ref{fig2.model}, the encoder uses 3D sparse convolution layers to account for the sparsity of 4D Radar point cloud data. Sparse convolution layers \cite{sparse_conv} are employed in stages where spatial resolution is reduced, while submanifold sparse convolution layers \cite{submaninfold} are utilized for operations where spatial resolution remains unchanged, thereby enhancing feature representation. In the decoder, 3D dense convolution layers are used to generate a dense 3D tensor. This method performs computations across all regions, making it suitable for producing complete tensors.

The generated tensor data is evaluated using a multi-scale discriminator  \cite{pix2pixhd}, which determines the authenticity of the data. To handle dense data, the discriminator also incorporates 3D dense convolution layers.

\subsection{Objective functions}
4DR P2T adopts a training framework using a Generator \(G\) and a Discriminator \(D\), inspired by traditional image translation methods. While image translation typically aims for a one-to-many mapping to generate diverse outputs, this study focuses on a one-to-one mapping, necessitating the design of appropriate loss functions. Following the method by Wang \cite{l2r}, the final loss function is defined as follows:
\begin{align}
\mathcal{L}(G, D) &= \mathcal{L}_{cGAN}(G, D) + \lambda_{L1} \mathcal{L}_{L1}(G) \nonumber \\
&\quad + \lambda_{perc} \mathcal{L}_{perc}(G) \label{eq:loss}
\end{align}
where \(\lambda_{L1}\) and \(\lambda_{perc}\) are weights that control the importance of each loss component, ensuring balanced training.

First, a conditional adversarial loss is used, where the \(G\) synthesizes data, and the \(D\) learns to distinguish between real and synthesized data. This is the core loss of cGANs, defined as:
\begin{align}
\mathcal{L}_{cGAN}(G, D) &= \mathbb{E}_{x,y}[\log D(x, y)] \nonumber \\
&\quad + \mathbb{E}_{x}[\log (1 - D(x, G(x)))] \label{eq:cgan_loss}
\end{align}
Here, \(x\) represents input data, \(y\) is the GT, and \(G(x)\) is the output of the \(G\). The \(D\) learns to differentiate \(G(x)\) from \(y\), while the \(G(x)\) is trained to deceive \(D\) by making \(G(x)\) resemble \(y\).

Second, L1 loss minimizes the absolute error between synthesized data \(G(x)\) and GT \(y\). This simple and stable loss function ensures that the synthesized data closely resembles real data:
\begin{equation}
\mathcal{L}_{L1}(G) = \mathbb{E}_{x,y}\left[|G(x) - y|_1\right]
\label{eq:l1_loss}
\end{equation}
Third, perceptual loss introduced in pix2pixHD is used to compare high-level feature distributions between synthesized and real data. By leveraging intermediate layer outputs from a pre-trained neural network, perceptual loss measures semantic differences, guiding the synthesized data to have similar high-level features to real data.

\begin{figure*}[!th]
{
  \centering
\vspace{1mm} 
 \includegraphics[width=1.0\textwidth]{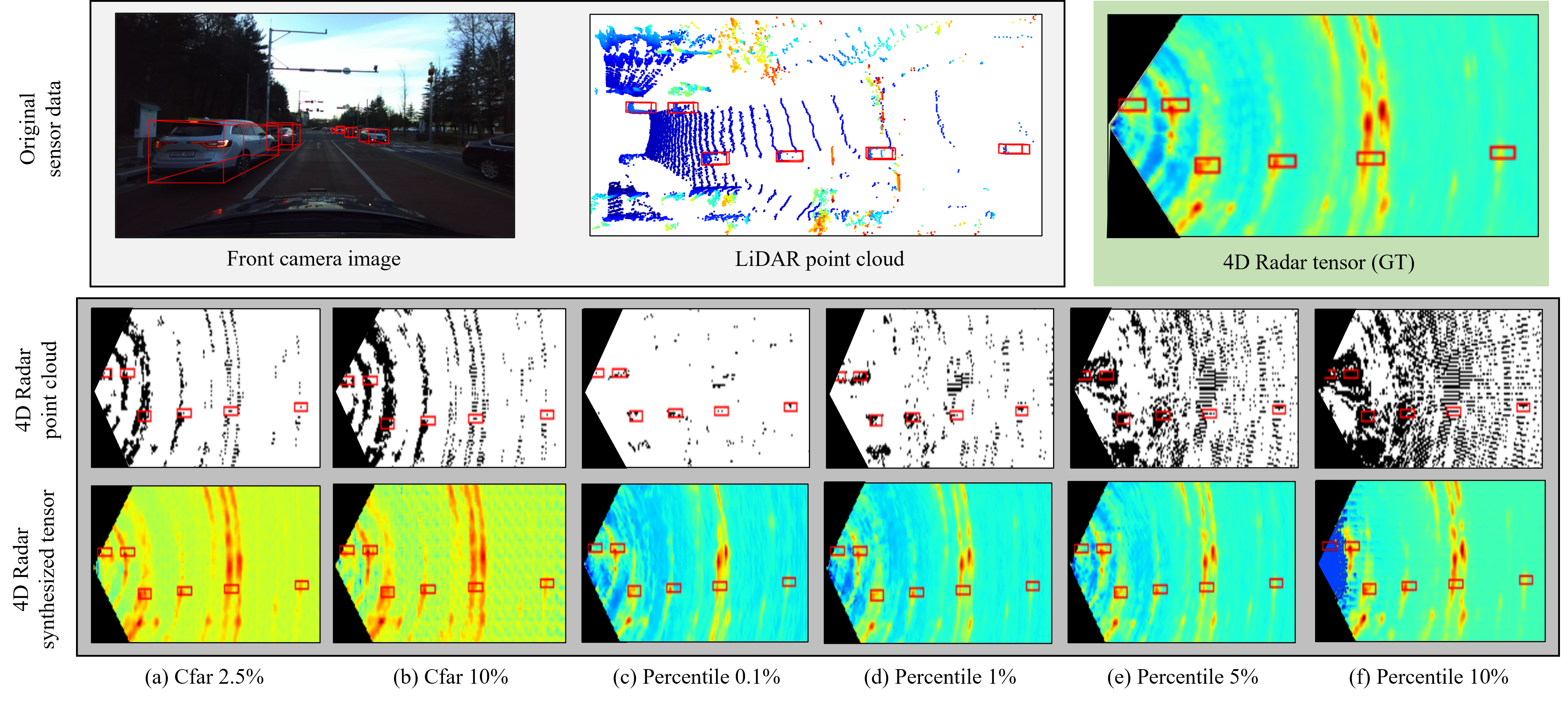}
    \caption{Qualitative experimental results of 4DR P2T. The top part shows the front camera image and LiDAR point cloud as reference data to understand the scene of the 4D Radar GT tensor data, while the bottom part presents the tensor data results generated by 4DR P2T under different point cloud generation methods and density conditions.}
  \label{fig3.result}
}
\end{figure*}

\section{Experiments} \label{sec:experiments}

This section describes the datasets and implementation details for training the 4DR P2T model, the evaluation metrics used, and the results and analysis of tensor generation performance.

\subsection{K-Radar dataset}
The K-Radar dataset \cite{KRadar}, which was used to train the 4DR P2T model, is the only dataset that provides 4D Radar tensor data (4DRT) consisting of the four dimensions: range, azimuth, elevation, and Doppler. This makes it of significant value. Additionally, K-Radar includes data from various weather conditions (clear, cloudy, fog, rain, sleet, light snow, and heavy snow), which distinguishes it from other autonomous driving 4D Radar datasets. Furthermore, the dataset includes 4D Radar data, high-resolution LiDAR data, and camera data, with 93.3K object labels for 35K frames, distributed across 58 different driving scenes.

K-Radar includes not only 4DRT tensor data but also point cloud data with CFAR applied, as used in the experiments of RTNH+ \cite{rtnh+}, and point cloud data with the percentile method applied, as used in the RTNH model \cite{KRadar}. The percentile method is effective in reducing memory and computational complexity while preserving the structure of tensor data, and thus was used as input data for training the RTNH model.

\subsection{Implementation details}
The experiments in this study were conducted using the K-Radar dataset, with data generated using various point cloud generation methods and density conditions for comparison. Specifically, point cloud data generated using the percentile method (top 0.1\%, 1\%, 5\%, 10\%) from enhanced K-Radar \cite{enhancekradar}, and point cloud data with hyper-parameter ($K_{1}$) of 2.5\% ($N_{2.5,a}$) and 10\% ($N_{10,a}$) using constant average CFAR from RTNH+ \cite{rtnh+} were used. These datasets were selected due to the significant differences in the point cloud distribution, making them suitable for comparison analysis.

The point cloud data used for training was extracted from 4D tensors in polar coordinates using CFAR or the percentile method and then converted into Cartesian coordinates. In this process, it can be observed that points become increasingly sparse as the range (distance) increases (Fig. \ref{fig0.overview}). The tensor data used for training was reconstructed into a dense cube shape through interpolation after converting from polar to Cartesian coordinates \cite{KRadar}. This data preparation process was set up to verify whether sparse point cloud data could be transformed into dense Cartesian tensor data and to expand its range of applicability.

The Region of Interest (ROI) was set as $x$-axis [0, 76.8], $y$-axis [-16, 16], and $z$-axis [-2, 10.8]. This range was chosen considering the scope of the RTNH\_WIDE \cite{kradargithub} object detection model trained with the widest range. All sequence data were used in the experiment, with the 4DR P2T model trained using the train set and performance evaluated using the test set. Model training was performed on an NVIDIA 3090 GPU, with a batch size of 8, a learning rate of 0.001, and Adam optimizer \cite{adam}, running for 20 epochs.

\subsection{Metrics}
For evaluation metrics, PSNR and SSIM were used, referencing \cite{l2r}. PSNR measures the signal-to-noise ratio between the synthetic data and the ground truth data, while SSIM measures the structural similarity between the two datasets. Both metrics indicate better performance with higher values. Although the generated data is a 3D tensor, the evaluation was performed by converting it to a 2D image through mean pooling along the height axis, and then calculating the metrics.

The deep-learning efficiency score (DES) metric, defined in Eq. \ref{eq:DES}, was used to identify efficient point cloud generation methods for deep learning. This metric aims to reduce data volume, which is related to point cloud density (PCD), while maintaining high tensor generation performance. First, the PSNR and SSIM values are normalized using min-max scaling, as shown in Eq. \ref{eq:psnr_norm} and Eq. \ref{eq:ssim_norm}, to ensure a fair comparison.
\begin{equation}
\text{PSNR}_{\text{norm}}^{(i)} = \frac{\text{PSNR}^{(i)} - \text{PSNR}_{\min}}{\text{PSNR}_{\max} - \text{PSNR}_{\min}}
\label{eq:psnr_norm}
\end{equation}
\begin{equation}
\text{SSIM}_{\text{norm}}^{(i)} = \frac{\text{SSIM}^{(i)} - \text{SSIM}_{\min}}{\text{SSIM}_{\max} - \text{SSIM}_{\min}}
\label{eq:ssim_norm}
\end{equation}
Where \( \text{PSNR}^{(i)} \) and \( \text{SSIM}^{(i)} \) represent the PSNR and SSIM for the \textit{\( i \)-th method}, respectively. \( \text{PSNR}_{\min} \) and \( \text{PSNR}_{\max} \) denote the minimum and maximum PSNR values across all evaluated methods, and similarly \( \text{SSIM}_{\min} \) and \( \text{SSIM}_{\max} \) represent the minimum and maximum SSIM values. Using these normalized values, the DES metric is computed as shown in Eq. \ref{eq:DES}.
\begin{equation}
M = \alpha \times \frac{\text{PSNR}_{\text{norm}}^{(i)}}{D^{(i)}} 
+ \beta \times \frac{\text{SSIM}_{\text{norm}}^{(i)}}{D^{(i)}}
\label{eq:DES}
\end{equation}
Where \( D^{(i)} \) is the PCD, defined as the ratio of detected points to the total possible points within the ROI for method \( i \). The weighting factors \( \alpha \) and \( \beta \), which control the relative importance of PSNR and SSIM, satisfy the constraint \( \alpha + \beta = 1 \). In this study, equal weights of 0.5 were assigned to both PSNR and SSIM.

\subsection{Results}

\begin{table}[ht]
\caption{Quantitative experimental results of 4DR P2T. 'Method' refers to the main categories of point cloud generation methods, while 'Hyper.' denotes the subcategories of point cloud generation methods, representing the hyper-parameters used in each point generation method. }
\label{tab:result}
\centering
\renewcommand{\arraystretch}{1.4} 
\begin{tabular}{c|c|c|c|c|c}
\hline \hline
\begin{tabular}[c]{@{}c@{}}Method\end{tabular} &
  Hyper. &
  \begin{tabular}[c]{@{}c@{}} PCD (\%)\end{tabular} &
  PSNR (dB) ↑ &
  SSIM ↑ &
  \begin{tabular}[c]{@{}c@{}}DES ↑\end{tabular} \\ \hline
\multirow{2}{*}{CFAR}       & 2.5 & 1.22   & 30.00 & 0.96 & 0.33 \\
                            & 10  & 2.42   & 28.14 & 0.96  & 0.11 \\ \hline
\multirow{4}{*}{Percentile} & 0.1 & 0.12   & 28.08 & 0.94 & 0.00 \\
                            & 1   & 1.11  & 31.66 & 0.96 & \textbf{0.48} \\
                            & 5   & 4.46 & \textbf{34.43} & \textbf{0.98} & 0.22 \\
                            & 10  & 8.17 & 30.00 & 0.96 & 0.05 \\ \hline \hline
\end{tabular}

\end{table}

Tab. \ref{tab:result} summarizes the tensor generation performance of the 4DR P2T model on 4D Radar point cloud data generated by various methods, evaluated using PSNR, SSIM, and DES. The average PSNR across all methods is 30.39dB—exceeding the 20–25 dB threshold commonly considered acceptable in wireless communication quality \cite{PSNR_1, PSNR_2}—indicating that the generated tensor data is of sufficiently high performance. As shown in Fig. \ref{fig3.result}, the percentile 5\% method achieves the best tensor generation performance, with a PSNR of 34.43dB and SSIM of 0.98. Meanwhile, the percentile 1\% method attains the highest DES value of 0.48, while also demonstrating superior point generation ability while reducing data volume, making it well-suited for deep learning model training.

\section{CONCLUSIONS} \label{sec:conclusion}

This study introduces the 4DR P2T model, which generates tensor data from 4D Radar point cloud data to address the limitation of inadequate spatial characteristic capture when CFAR is applied to 4D Radar data. By leveraging a cGAN-based architecture, our model effectively generates tensor data, as demonstrated by an average PSNR of 30.39dB and SSIM of 0.96. In addition, our comparative experiments show that the percentile 5\% method yields the best tensor generation performance,  while the percentile 1\% method offers an optimal balance between data volume reduction and performance, making it well-suited for deep learning training.

Future research will extend the 4DR P2T model to accommodate unpaired data, enabling tensor generation even for datasets lacking original tensor data.  Additionally, Doppler information will be incorporated to further enhance object representation. These advancements aim to improve the preservation of critical object features, enhance sensor fusion, and ultimately strengthen perception capabilities in autonomous driving systems.

\addtolength{\textheight}{-12cm}   



\section*{ACKNOWLEDGMENT}

This work was supported by the National Research Foundation of Korea (NRF) grant funded by the Korea government (MSIT) (No. 2021R1A2C3008370).


\bibliographystyle{unsrt}
\bibliography{ref}

\end{document}